\documentclass{midl} 
\def\blind#1{#1}

\usepackage[symbol]{footmisc}
\usepackage{booktabs}

\jmlryear{}
\jmlrproceedings{}{}
\jmlrvolume{}
\editors{}
\jmlrpages{}

\title[Explaining 3D Computed Tomography Classifiers with Counterfactuals]{Explaining 3D Computed Tomography Classifiers \\with Counterfactuals}

\midlauthor{\Name{Joseph Paul Cohen}\textsuperscript{$\dagger$} \Email{ joseph@josephpcohen.com}\\
\addr Stanford University
\AND
\Name{Louis Blankemeier}\\
\addr Stanford University
\AND
\Name{Akshay Chaudhari}\\
\addr Stanford University
}

\begin{document}

\maketitle

\footnotetext{\textsuperscript{$\dagger$}Work not related to position at Amazon.}

\begin{abstract}
Counterfactual explanations enhance the interpretability of deep learning models in medical imaging, yet adapting them to 3D CT scans poses challenges due to volumetric complexity and resource demands.
We extend the Latent Shift counterfactual generation method from 2D applications to explain 3D computed tomography (CT) scans classifiers. 
We address the challenges associated with 3D classifiers, such as limited training samples and high memory demands, by implementing a slice-based autoencoder and gradient blocking except for specific chunks of slices. 
This method leverages a 2D encoder trained on CT slices, which are subsequently combined to maintain 3D context. 
We demonstrate this technique on two models for clinical phenotype prediction and lung segmentation. 
Our approach is both memory-efficient and effective for generating interpretable counterfactuals in high-resolution 3D medical imaging.
\end{abstract}

\section{Introduction}

Neural Networks can learn to identify features and make predictions from computed tomography (CT) scans predicting clinical phenotypes \citep{Blankemeier2024Merlin} and survival \citep{Thanoon2023CTSurvival}. 
To understand why these predictions are made, it is crucial to have interpretable explanations, especially in the medical domain where high-stakes decisions are made. 
Counterfactual (CF) images, are synthetic images that simulate a change in the class label of an input image with respect to what features a classifier is using to make a prediction \citep{Wachter2017BlackBoxGDPR}. 
The counterfactuals we generate do not correspond to human-defined concepts or gold-standard labels but instead reflect the classifier's decision-making process. These counterfactuals can then be used to audit the features used by the classifier.

In this work we extend the counterfactual generation method of Latent Shift, previously demonstrated for 2D counterfactuals (CFs) \citep{Cohen2021gif}, into the 3D domain to enable it to work for 3D classifiers of entire CT volumes.
While a natural approach would be to use a 3D autoencoder, this presents difficulties such as the scarcity of training data for 3D autoencoders and the significant computational and memory demands, especially for large models and high-resolution data, both of which are likely required for generating high-quality counterfactuals.

To address this, we employ a 2D autoencoder by slicing the 3D volumes, then encoding, decoding, and concatenating the slices. 
This slice-based approach is more data-efficient than training directly on entire volumes, helping to mitigate overfitting by avoiding biases related to specific acquisition views (e.g., abdomen or head). 
By randomly sampling slices, the model is less likely to learn unwanted correlations connected to particular views.

Challenges posed by memory constraints on GPUs persist, particularly when calculating a partial derivative with respect to the latent space, preventing counterfactual generation for entire volumes. 
To address this the gradient is propagated for only a subset of the slices which limits where in the volume the changes can be made. 
Our findings indicate that this approach minimally affects the effectiveness of the generated counterfactuals.

Contributions:
\begin{itemize}
\setlength\itemsep{0.09em}
    \item Train and make public a VQ-GAN on 1,447,551 slices from 12,686 CT scans to enable CF generation.
    \item Develop the first approach to generate CFs of 3D CT classifiers.
    \item Study two prediction models: one for clinical phenotype prediction and another for lung segmentation.
\end{itemize}

\section{Counterfactual Generation}

The Latent Shift method \citep{Cohen2021gif} is employed to generate counterfactuals (CFs). 
In order to make modifications to the image this method requires computing the gradient from the output of the classifier to the latent space of an autoencoder $$\frac{\partial f(D(E(x)))}{ \partial z} \text{ \hspace{20pt}where \hspace{20pt}} z = E(x).$$ 
The autoencoder serves as a dimensionality reduction tool, mapping the high-dimensional input image into a compact latent space where meaningful transformations can be applied. 
$E$ and $D$ are an encoder/decoder pair trained to represent the domain of CT Scans (see \S\ref{sec:ae}). 
Using this trained decoder ensures that the generated counterfactuals remain within a plausible distribution of medical images, reducing the risk of unrealistic modifications.
This gradient is then subtracted from the latent representation of the input image. 
By subtracting this gradient, scaled by a coefficient $\lambda$, from the latent representation, we aim to modify the image in a direction that decreases the classifier's confidence in its original prediction. 
The coefficient $\lambda$ is selected using an iterative search to ensure a predetermined change in the classifier's output.

\begin{figure}[t]
    \centering
    \vspace{-10pt}
    \includegraphics[width=0.95\linewidth]{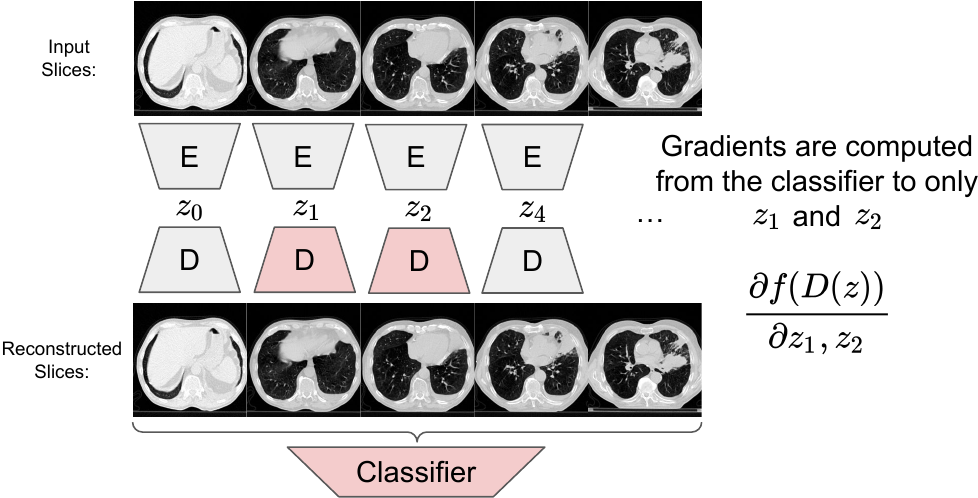}
    \caption{Illustration of the slice-based autoencoder (Slice AE) approach for generating counterfactuals in 3D CT volumes. Only selected latent representations, $z_1$ and $z_2$, have gradients computed from the classifier output, enabling memory-efficient counterfactual generation while preserving the ability to navigate and modify specific parts of the 3D volume.}
    \label{fig:SliceAE}
\end{figure}

Computing this gradient is memory-intensive, particularly when using a 3D classifier and autoencoder. 
In order to make this tractable, a 2D autoencoder is used which allows us to propagate the gradient to only a subset of slices, and their respective latent embeddings. 
We call this approach Slice AE and depict it in Figure \ref{fig:SliceAE}. 
With this approach the entire volume is encoded slice by slice and then decoded and concatenated into a full volume and input into the classifier. 
During decoding, gradients are computed for only a subset of slices.
The remaining slices have the gradient blocked. 
This approach allows the classifier to compute gradients for a few slices without sacrificing the context of the entire volume. 
Typically only 10 slices can have their gradient computed using 32GB of memory on a Tesla V100 used for this work. For one chunk (e.g. 10 slices) it takes 30 seconds to compute the CF image.

This gradient can then be used to modify the latent representation and then decoded and concatenated and input into the classifier to determine if the the prediction has been sufficiently reduced. 
An iterative search is used by reducing the value of $\lambda$ until the prediction stops decreasing. This search process will also stop if the total change in pixel values is more than 5\% which typically represents a failure in generating the CF. 
One challenge that still can remain is if the relationship between neighboring slices is not preserved, for this initial approach no special regularization is performed as it was not observed to be a major issue, but it would likely improve the consistency of the volume.

\section{Model and Training}
\label{sec:ae}

We train a VQ-GAN \citep{Esser2020TamingTransformers} which is a VQ-VAE (vector quantized variational autoencoder) with a perceptual adversarial loss. This approach generates high-resolution images while having an encoder/decoder which provides a latent space that can be navigated using CF generation methods. This is in contrast to diffusion models which also generate high-resolution images but the latent space requires iterations, so it is not trivially compatible with existing CF generation methods.

This model is used to reconstruct at the slice level instead of the entire volumes. This is done for data efficiency as volumes are very high dimensional and don't always capture the same extent, which would likely lead to overfitting. To further maximize data efficiency, images are scaled down by 1/2 (often 512x512 $\rightarrow$ 256x256) and the model is trained on crops of 128x128. Training on crops enables the use of volumes that have already been cropped from a larger volume, which is commonly found in public datasets. This allows the model to observe multiple organs at once in the receptive field. The model was trained on a single 16GB P100 for 386 epochs which took 810 hours (33 days) of GPU compute time.
Code and model weights in PyTorch \citep{Paszke2019pytorch} are available online\footnote{Code and model weights available: \href{\blind{https://github.com/ieee8023/ct-counterfactuals}}{\blind{https://github.com/ieee8023/ct-counterfactuals}}}.

\section{Data}

The study utilizes three datasets selected for their size, public availability, coverage of various organ systems, and inclusion of both healthy and diseased states. The LUNA16 dataset \cite{Setio2016LUNA16} contains imaging data focused on the lungs and includes cancerous lesions. A total of 227,225 slices were obtained from 888 volumetric scans. The TotalSegmenter dataset \cite{Wasserthal2022TotalSegmentator} provides imaging of all major organs, with 312,400 slices obtained from 1,204 scans. The DeepLesion dataset \cite{Yan2017DeepLesion} includes images of multiple organs, both healthy and with lesions, offering a robust representation of diseased tissues. From 10,594 scans, 907,926 slices are obtained. Collectively, these datasets ensure comprehensive representation of both healthy and unhealthy organs, contributing to providing a good data representation to the model.

\section{Experiments}

\subsection{Lung Size}

\begin{figure}[t]
    \centering
    \includegraphics[width=1\linewidth]{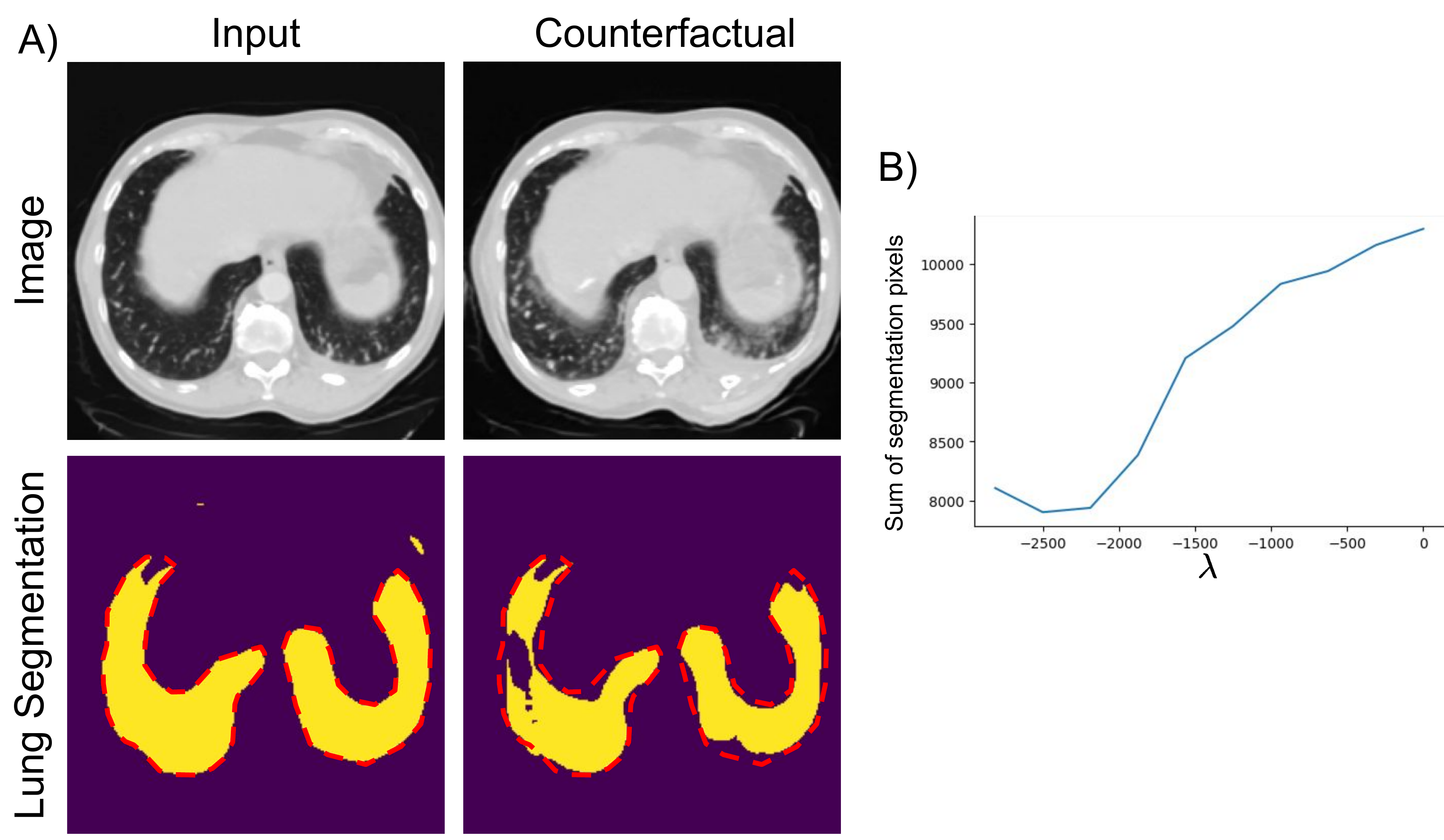}
    \caption{A) An example CF generated for lung size. The model's predicted segmentation mask for lung is shown below. A red outline tracing the segmentation on the input image is overlaid in the segmentation of the CF image as well, confirming the reduction in predicted lung size. B) A plot showing the sum of pixels predicted as lung as the $\lambda$ is changed during the Latent Shift CF generation process.}
    \label{fig:lungsize}
\end{figure}

In order to verify that this method works, an easily auditable CF can be generated for lung size. The work \cite{Hofmanninger2020CTLungSeg} released a lung segmentation model that we can transform into a regression task by taking the sum of the segmentation outputs for the lung class. As more pixels are classified as lung, the larger the sum will become. Figure \ref{fig:lungsize}A visualizes the CF generated for this task, demonstrating a strong visual signal that the lung size is reduced. Using that same model to segment the image and CF image, the size of the segmentation has been reduced. Here the segmentation is thresholded to 0.5 to make the comparison easier.

The reduction in lung size can also be validated by looking at the sum of the lung pixels (Figure \ref{fig:lungsize}B). We should observe that lowering the $\lambda$ value reduces the total sum of pixels, which is what is observed.
We do not observe an infinite monotonic reduction because the CF generation process is limited by the latent variable model's ability to remove these features along the specific vector in the latent space identified by Latent Shift. Although it is more likely that a lung size smaller than what we achieve is outside the domain of the latent variable model and not something it can represent.

\subsection{Plural Effusion}

Plural effusion, characterized by the accumulation of fluid between the pleural layers surrounding the lungs \citep{Krishna2025PleuralEffusion}, can be visually identified in CT scans as bright regions along the lung periphery. To validate that the classifier utilized in \cite{Blankemeier2024Merlin} is using the correct features, we applied our counterfactual generation method to this task.

\begin{figure}
    \centering
    \includegraphics[width=1\linewidth]{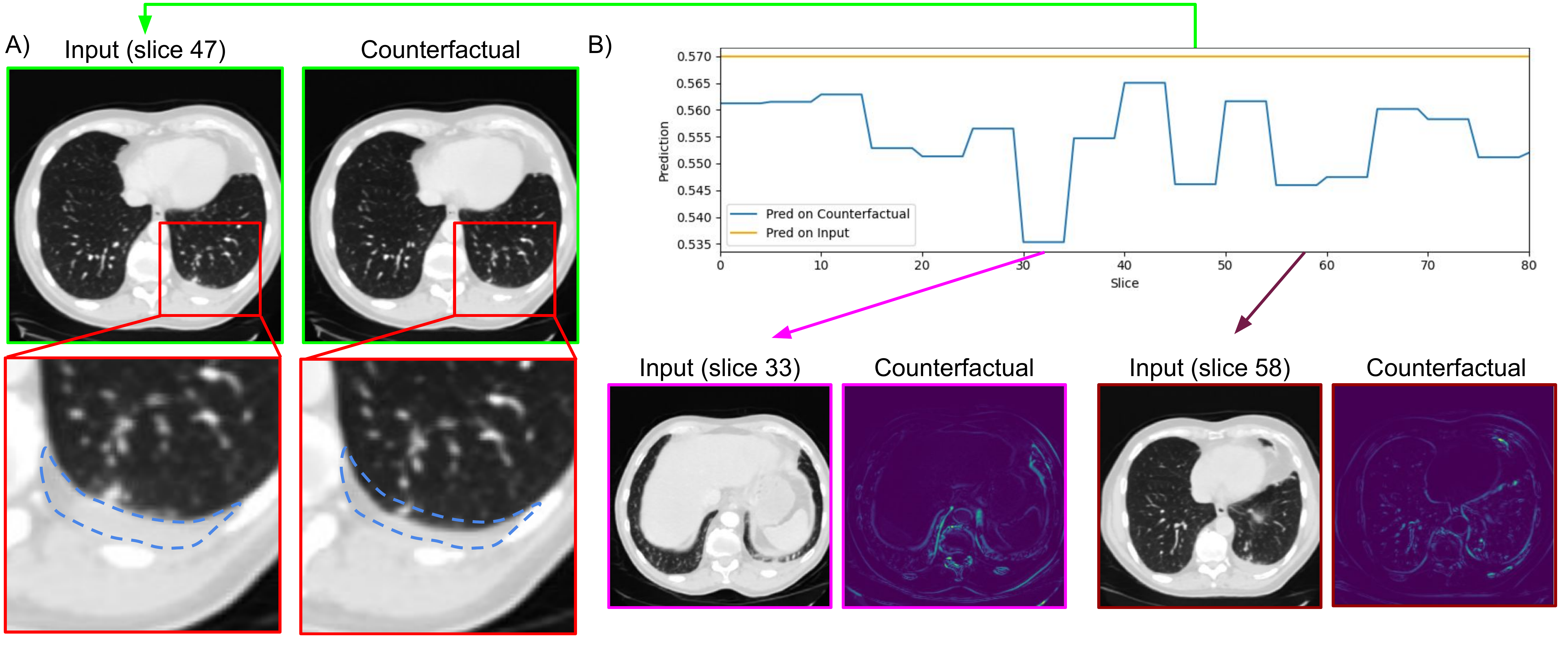}
    \caption{A: The input and CF slices showing a reduction in plural effusion. The blue dashed line outlines the side of the lung area that is reduced in the CF. B: Localization of CF slices that contribute to changes in the prediction. The volume is processed in chunks of five slices, restricting changes to that region. Slices between 30 and 35 are identified as producing the most change while slices 45-50 (Shown in A) and 55-60 also reduce the prediction. Heatmaps of these changes between the input and CF are shown.}
    \label{fig:explain}
\end{figure}

Figure \ref{fig:explain}A demonstrates the input and counterfactual (CF) slices, particularly at indices 33, 47, and 58. The sample LUNG1-001 is used from the of the NSCLC-Radiomics dataset \cite{Aerts2014NSCLC-Radiomics}. In these CF slices, reductions in the pleural effusion regions are visibly evident, outlined by dashed blue lines. This confirms that the generated counterfactuals correctly target the bright regions corresponding to fluid buildup, providing evidence that the classifier relies on appropriate features for its predictions.

To further analyze the classifier's decision-making process, we localized the slices that contributed most significantly to changes in the prediction. Figure \ref{fig:explain}B shows heatmaps highlighting areas within slices where the differences between the input and CF are most pronounced. By processing the volume in chunks of five slices and restricting changes to specific regions, we identified slices 30-35, 45-50, and 55-60 as having the greatest impact on the prediction. For instance, slices 45-50, visualized in Figure \ref{fig:explain}A, exhibit substantial reduction in the pleural effusion region, aligning with a decrease in the classifier's confidence.

\subsubsection{Saliency Comparison}

\begin{figure}
    \centering
    \includegraphics[width=1\linewidth]{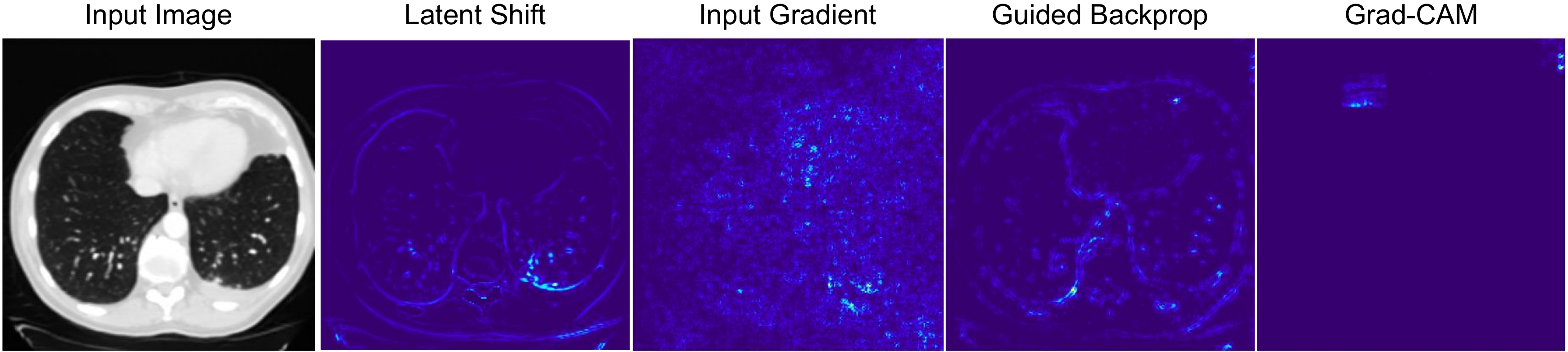}
    \caption{A comparison with other explanation methods which use gradient influence to highlight relevant pixels. Slice 45 is visualized from LUNG1-001 of the NSCLC-Radiomics dataset. }
    \label{fig:saliency}
\end{figure}

To evaluate the effectiveness of existing explainability methods we compare against classic baselines implemented in the Captum library in Figure \ref{fig:saliency}. Input Gradient \cite{Simonyan2014}, Guided Backprop \cite{Springenberg2014GuidedBackprop}, and Grad-CAM \cite{Selvaraju2017GradCAM}. The slice selected has a clear effusion present (circled in Figure \ref{fig:explain}) which is identifyed by Latent Shift as well as the Input Gradients. If we make the assumption that the classifier is correctly using this feature then the other methods fail, otherwise these results are conflicting.

\subsubsection{Quantitative Evaluation}

To quantitatively assess the impact of CF images on the classifier, we measure the change in prediction across a dataset. The literature uses the term ``prediction gain'' \cite{Nemirovsky2020CounteRGAN}, but we focus on evaluating prediction reduction.

We perform this evaluation for the pleural effusion task using the PleThora dataset \cite{Kiser2020PleThora}, which contains labels for pleural effusion from the public NSCLC-Radiomics dataset \cite{Aerts2014NSCLC-Radiomics}. This dataset contains 402 CT volumes where 78 are labeled positive for plural effusion. Due to a mismatch between data and labels, we evaluated 399 CT volumes, of which 77 had positive labels. No prior work, to our knowledge, generates counterfactuals for CT scans to serve as a baseline for evaluation. A maximum chunk size of 12 was adopted because larger sizes exceeded memory capacity for certain volumes.

To compute a counterfactual for a volume using the latent shift approach, only a chunk of slices is modified at a time, resulting in a reduced prediction for the entire volume. We report the minimum prediction here. Computation time for an entire volume is determined by dividing the number of slices by the chunk size and multiplying by the per-chunk processing time. For a 100-slice volume with a chunk size of 10, this requires 300 seconds.
Alternatively, combining CF chunks into a single composite volume introduces inconsistencies, as each chunk is computed independently. The chunk-by-chunk approach provides locality (see Figure \ref{fig:explain}) by identifying regions that, when altered, reduce the prediction.

\begin{figure}
    \centering
    \includegraphics[width=1\linewidth]{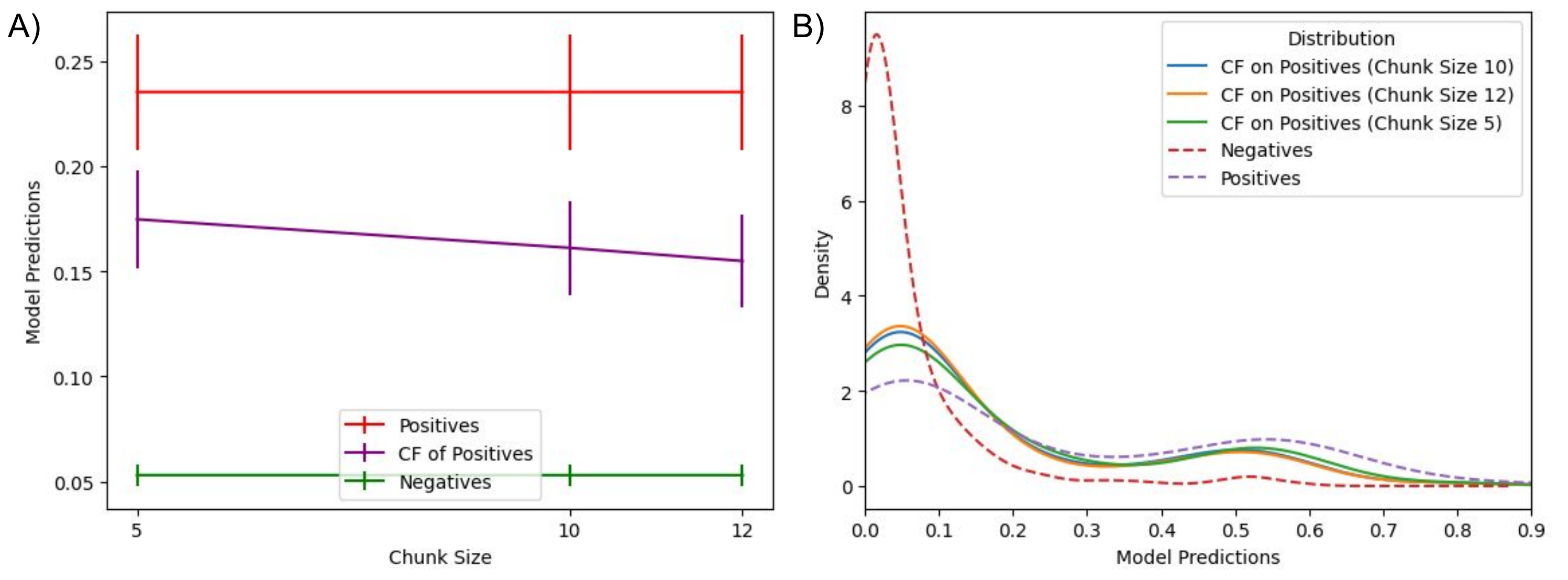}
    \caption{A) Results varying the chunk size when computing CFs on the examples. Error bars are standard error. B) The change in distribution of model predictions between the positive/negative examples and the CFs of positive examples.}
    \label{fig:pe_predgain_plots}
\end{figure}

In Figure \ref{fig:pe_predgain_plots}A, we vary chunk size during CF computation on the examples. A larger chunk size appears to offer a larger decrease in model prediction, at the cost of reduced localization in the scan. Figure \ref{fig:pe_predgain_plots}B illustrates the change in distribution of model predictions between the positive/negative examples and the CFs of positive examples. The CF volumes yield predictions more aligned with the distribution of negative examples.

Table \ref{tab:pe_numbers} shows a 17-point difference between positive and negative examples. We also observe when our CF method is applied to the positive examples the prediction is reduced by 8 points, indicating an effective counterfactual generation method. We also observe a reduction in prediction when our CF method is applied to negative examples. The difference between positive and negative examples is statistically significant ($p < 0.001$).

\begin{table}[]
    \centering
    \begin{tabular}{c c}
        \toprule
        Label & Model Prediction \\
        \midrule
        Positive Examples (N=77) & 0.24$\pm$0.03\\
        Negative Examples (N=322) & 0.05$\pm$0.01\\
        CFs for Positive Examples (N=77) & 0.15$\pm$0.02\\
        \bottomrule
    \end{tabular}
    \caption{Model predictions made on input data and CFs of input data. The mean and standard error are shown. A chunk size of 12 is used. }
    \label{tab:pe_numbers}
\end{table}

\section{Conclusion}

This work is the first to generate counterfactual images for 3D CT classifiers.
Our approach mitigates the challenges of limited training data and high memory demands that are needed to apply the Latent Shift approach to this domain. The CT latent variable model used to achieve this is made publicly available as it utilized significant computational resources (810 GPU hours).

Case studies confirm that the generated counterfactuals target clinically relevant features, underscoring this method’s potential to enhance the transparency and trustworthiness of AI systems in medical applications.
Furthermore, our approach enables localized analysis, offering insights into the regions most influential in the classifier’s predictions. This capability highlights the potential of counterfactual explanations to enhance transparency and trust in AI systems, particularly in high-stakes medical applications.

\section{Related Work}

In medical imaging, where trust and accountability are paramount, counterfactuals have been applied to 2D radiology modalities such as chest X-ray \cite{Singla2021BlackBoxSmoothly, Seah2019GVR_CFs}. These employed counterfactuals to explain classifications, pinpointing regions critical for diagnosing conditions like pneumonia and plural effusion.

Despite these advances, applying counterfactual explanations to 3D imaging modalities like CT scans remains largely unexplored, inspiring our work to address this gap.
Existing work on 3D MRIs employed 2D slice based CF generation using causal modeling \cite{Ribeiro2023SCM_CF} and conditional generation \cite{Kumar2022MRICF}. 
Work by \citet{Peng2024BrainMRI3D} focused on 3D MRI CF generation, also using a VQ-GAN but combined with causal modeling to generate the CF volumes. 

These works rely on conditionally generating CF images rather than directly explaining an arbitrary 3D classification model. The generated images may not align with the classifiers we aim to explain, which is why we chose the Latent Shift approach, as it modifies image features based directly on the classifier. This also allows arbitrary classifiers to be explained using the same latent variable model used in our research.

\section{Limitations}

Potential Domain Bias: The datasets used for training the autoencoder primarily include publicly available medical scans, which may not fully represent the diversity of real-world clinical data. This could impact generalizability and robustness when applied to new datasets or different scanning protocols.

Lack of Clinical Validation: While the generated counterfactuals visually align with expected anatomical changes, we have not conducted a formal clinical validation study to assess their interpretability and usefulness for medical practitioners in clinical decision-making.

Identity Preservation/unintended distortions: Counterfactual explanations should only change features used by the classifier. The proposed approach follows the gradients of the classifier regularized by a decoder which does not provide a guarantee, but has been empirically shown to work. This is challenging with the proposed approach due to bias introduced by the latent variable model as some features may not be able to be modulated or are entangled with unrelated features.

\section{Acknowledgments}

\blind{We would like to thank Stanford University and the Stanford Research Computing Center for providing computational resources and support that contributed to these research results.}

\bibliography{refs}

\end{document}